\documentclass[9pt]{extarticle}
\usepackage[T1]{fontenc}

\usepackage[table,xcdraw]{xcolor}

\usepackage{spconf,amsmath,graphicx,tikz,amssymb,amsfonts,bm,mathtools,mathabx}
\usepackage{multicol}
\usepackage{amsthm}
\usepackage{float}
\usepackage{comment}

\usepackage[linkcolor=mar,colorlinks=true]{hyperref}
\hypersetup{colorlinks,breaklinks,
            linkcolor=[rgb]{0.5,0,0},
            citecolor=[rgb]{0.39,0.7,1.0}}
\usepackage{cleveref}
\usepackage{cite}
\usepackage{enumitem}
\usepackage{outlines}
\usepackage{algpseudocode}
\usepackage{algorithm}

\usepackage[skip=0pt]{caption}
\usepackage[skip=0pt]{subcaption}
\usepackage{bm}

\newtheoremstyle{reduced}
  {6pt} 
  {6pt} 
  {} 
  {} 
  {\bfseries} 
  {.} 
  {.5em} 
  {} 

\theoremstyle{reduced}

\usepackage{adjustbox}



\renewenvironment{thebibliography}[1]{%
   \begin{oldthebibliography}{#1}%
     \setlength{\itemsep}{0ex}%
}%
{%
   \end{oldthebibliography}%
}
\title{Federated Learning Under Restricted User Availability\vspace{-10pt}}
\name{Periklis Theodoropoulos, Konstantinos E. Nikolakakis, Dionysis Kalogerias \vspace{-10pt}}
\address{Department of EE---Yale University\\ 
 \small{\tt \{\href{mailto:periklis.theodoropoulos@yale.edu}{periklis.theodoropoulos}, \href{mailto:konstantinos.nikolakakis@yale.edu}{konstantinos.nikolakakis},
 \href{mailto:dionysis.kalogerias@yale.edu}{dionysis.kalogerias}\}@yale.edu}
\vspace{-14pt} }

\begin{document}
\setlist[itemize]{noitemsep, topsep=1pt}
\setlist[itemize]{leftmargin=*}
\maketitle
\begin{abstract}
\vspace{-3pt}
Federated Learning (FL) is a decentralized machine learning framework that enables collaborative model training while respecting data privacy. In various applications, non-uniform availability or participation of users is unavoidable due to an adverse or stochastic environment, the latter often being uncontrollable during learning. Here, we posit a generic user selection mechanism implementing a possibly randomized, stationary selection policy, suggestively termed as a Random Access Model (RAM). We propose a new formulation of the FL problem which effectively captures and mitigates limited participation of data originating from infrequent, or restricted users, at the presence of a RAM. By employing the Conditional Value-at-Risk (CVaR) over the (unknown) RAM distribution, we extend the expected loss FL objective to a risk-aware objective, enabling the design of an efficient training algorithm that is completely oblivious to the RAM, and with essentially identical complexity as FedAvg. Our experiments on synthetic and benchmark datasets show that the proposed approach achieves significantly improved performance as compared with standard FL, under a variety of setups. 

\end{abstract}\vspace{-5pt}

\vspace{-11pt}
\section{Introduction} \label{sec:intro}
\vspace{-9pt}

Federated learning (FL) is a distributed learning framework allowing multiple users to train a global model collaboratively without sharing their local data \cite{smith2017federated}. In recent years, the classical Federated Averaging approach (FedAvg) has developed into an essential learning paradigm \cite{mcmahan2017communication}, following a certain basic workflow where each user locally updates its own model parameters and then periodically sends its updated parameters to a central server. The server then appropriately aggregates the received updates of the users and broadcasts to them the new global model parameters. This scheme is repeated until the global model converges \cite{kairouz2021advances}.



Vanilla FedAvg faces the problem of instability, caused either due to non- indepedent and identically distributed (non-iid) data among users (known as ''client-drift'')~\cite{karimireddy2020scaffold,acar2021federated,li2020federated}, or due to non-uniform user availability, which happens as a result of several factors, such as network outages \cite{wang2021field}, battery drain \cite{eichner2019semi}, or user inactivity \cite{ruan2021towards}. For instance, when a user device is unavailable, it cannot communicate with the server for a certain time interval. This leads to incomplete global model convergence, to a possible bias of the global model towards the data from the most available users, and less accurate performance of the global model to the data that the less available user have, especially in a strongly heterogeneous regime.

In prior works, several methods address the challenge of non-uniform client availability in FL. For the (server) aggregation step, a set of techniques focuses on re-weighting the model parameters of the users by adapting their weights dynamically. In particular, some works focus on the adaptation of the aggregation coefficients based on several criteria, such as completion of the local steps the users have done \cite{ruan2021towards}, or the temporal correlation and their low availability \cite{rodio2023federated}, or their conformity level \cite{laguel2021superquantile}, or their performance on previous steps \cite{zhao2022dynamic}.
Moreover, the non-uniform availability of users could be treated by dividing users into blocks based on their relative frequency, and applying a pluralistic aggregation step at each block has been proposed in \cite{eichner2019semi}. Lastly,  \cite{da2022multichannel} suggests a different FL system that adopts a multiple-channel approach, following specifically the ALOHA protocol and adapting the access probability of users based on their local updates. 

A second group of techniques explores optimal user sampling strategies. Selection of users based on local characteristics, such as their local performance, level of importance, or irrelevance, is proposed in \cite{rai2022client,chen2020optimal,cho2020client}, respectively. 
Optimal user selection has also been explored by minimizing the variance of the numbers of times the server samples a client \cite{wang2023fedgs}, or by learning a selection strategy for clients with intermittent availability  \cite{ribero2022federated}. Further, \cite{tang2023tackling} uses stratified user sampling based on their data statistics to address system-induced bias under time-varying client availability. Lastly, a well-defined communication protocol where the server periodically selects user devices that meet appropriate criteria, is proposed in \cite{bonawitz2019towards}.

\begin{figure}[!t]
  \centering  \centerline{\includegraphics[width=0.445\textwidth]{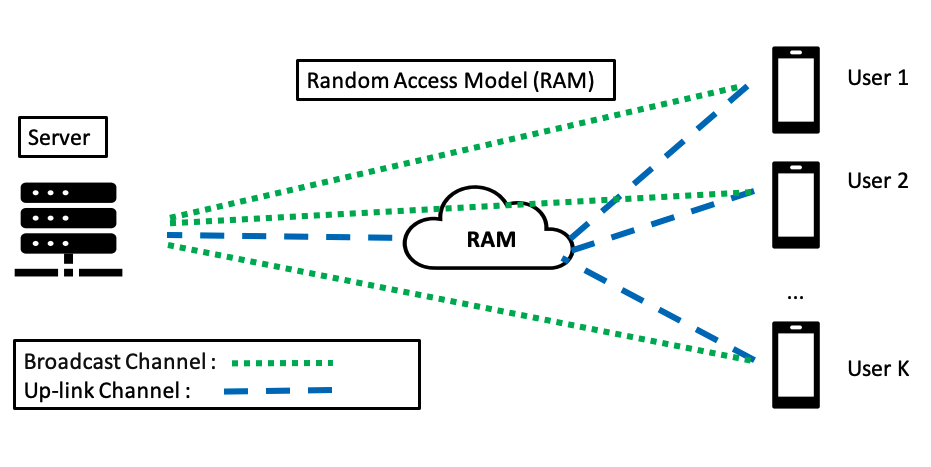}}
\vspace{-5bp}
\caption{Users send their updates to the RAM, which relays the data of only one user to the server. The server broadcasts the received local model to all users, and the process repeats.
}
\label{fig:RAM process}
\end{figure}


\textit{In this work}, we deal with two key limitations, critical in practical scenarios. First, we consider a stochastic environment allowing only one (for simplicity) user to transmit data each time, with different and possibly highly biased selection probabilities (weights) for each user (restricted random user selection). Secondly, the server is completely agnostic to user selection probabilities, and cannot direct user participation. Under these conditions, the server communicates with the users through a noisy channel potentially adversarial to the learning procedure. We posit an intermediate provider, e.g., a multiplexer or switch, which we call the \textit{Random Access Model (RAM)}, between users and the server. The RAM is responsible for relaying user model updates to the server by selecting a certain number of users to relay at each update (and communication) round.

The user selection criteria of the RAM remain unknown to both the server and users. Therefore, at least from the perspective of the server and users, the simplest approach to describe RAM user selection is by using a memoryless probabilistic model, i.e., the RAM spits out each user with some fixed probability, independently across communication rounds.  In other words, the RAM acts as a stationary \textit{erasure channel}, where at each round the updates of only one user survive, while the updates of the rest of the users are discarded, according to a fixed user selection distribution. That is, the RAM implements a possibly randomized user selection policy.  
Fig. \ref{fig:RAM process} depicts this scheme, where the RAM allows only one user to relay its local model to the server each time. The server cannot intervene and works as a simple broadcaster. 


Under this setting, we propose a FL approach which is agnostic to user selection implemented by the RAM, by extending the expected empirical loss to a weighted risk-aware objective, using the \textit{Conditional Value-at-Risk} (CVaR) over the (unknown) distribution dictating such selection of users. This approach is, to the best of our knowledge, new and results in a robust training algorithm 
which exploits the structure of FedAvg, with \textit{at most} identical computational, iteration and communication complexity, at the expense of tuning two additional hyper-parameters. Our approach demonstrates notably superior performance compared with standard FL, which struggles to achieve accurate data classification when operating under limited access to user updates. Our experimental evaluations take place on both synthetic and standard benchmark datasets.

As a motivating example we consider a toy logistic regression problem on 2D synthetic data in Fig. \ref{fig:Simple Logistic Regression Example for 2D data}. Each user trains on distinct patterns; this setup may resemble, e.g., sensors at different locations, each observing features corresponding to those patterns. Even for an easily verifiable setting with three or four classes, we observe that standard FL fails to classify data from less frequent users, while the proposed approach succeeds at finding decision boundaries that correctly classify the data from all users. All experiments were performed under the same training conditions (number of global rounds, step sizes, etc.). This example is elaborated in detail in Section \ref{Simple motivating example}.
\color{black}

\vspace{-14pt}
\subsection{Applicability of the RAM}\label{Concrete Applications}
\vspace{-7pt}
The RAM provides an abstraction for capturing intricacies involved in the communication among network nodes in numerous networking applications. In the following, we discuss some relevant examples. More specifically, user participation with different relative frequencies naturally appears in the context of routing or switching. In this case, the RAM may embody a router or switch operating at the network layer, implementing some possibly randomized (steady-state) device selection policy, with the goal of routing numerous requests intelligently.
The RAM may also model the role of the MAC sub-layer of a data-link network layer in
managing access to a shared communication medium and 
potentially optimizing traffic prioritization for multiple users by, e.g., performing informed bandwidth allocation to different devices. Further, the RAM could describe potential user unavailability as an impact of network infrastructure at the physical layer. In this case, the RAM models the effects of varying traffic or interference patterns, resulting from allocating physical resources, such as power or frequency (carrier) directly.

In all those setups, highly biased and non-uniform user selections may result naturally under common circumstances. These include, for instance, service provider imbalanced user priorities due to certain contracts or monetary constraints, geographical constraints (e.g., communication in rural areas or in underwater applications),
frequency band availability or scarcity in the physical layer, and opportunistic resource allocation policies. Additionally, user selection bias may be induced as a result of network outages and disruptions caused by environmental conditions, equipment failures, power outages, or even natural disasters, affecting user network access dramatically and thus causing non-uniformity in the availability of user data. Last but not least, user selection bias may result due to communication rate limitations caused by privacy concerns and censoring related to data from certain sources or of certain types.

The RAM provides a convenient abstraction for modeling such non-uniform user selection schemes, which introduce unavoidable data scarcity and result in rare user participation in a FL setup.
\vspace{-5pt}

\vspace{-6pt} 
\subsection{Comparison with previous works}\label{Comparison with previous works}
\vspace{-6pt}



Due to rare user participation enforced by the RAM, together with its unknown structure, existing techniques (e.g., user re-weighting \cite{ruan2021towards,rodio2023federated, laguel2021superquantile, zhao2022dynamic} and optimal sampling policies \cite{rai2022client,chen2020optimal,cho2020client, wang2023fedgs, ribero2022federated, tang2023tackling}) are inapplicable, as they require server-user interaction. In contrast, the proposed algorithm allows the server to manage user selection bias by optimizing efficiently under data scarcity. Our algorithm can in fact systematically manage non-uniform user participation, by adaptively focusing more on less frequent users. This is possible by exploiting native properties of the CVaR. The proposed technical approach also shares some similarities with \cite{li2019fair, zhang2022proportional}. However, our formulation is fully interpretable (i.e., all hyperparameters admit a fully specified operational meaning), while the resulting algorithm operates fully within the framework of FedAvg; no heuristics or customized algorithm design is necessary.
Lastly, our approach resembles online federated learning \cite{mitra2021online} but differs in that users can only access the server when permitted by the RAM, even after completing local updates.

\vspace{-9pt}
\section{Problem Setup} \label{Problem setup}
\vspace{-8pt}

\begin{figure}[!t]
     \centering
     \begin{subfigure}[b]{0.45\textwidth}
        \centering
        \includegraphics[width=0.3\textwidth]{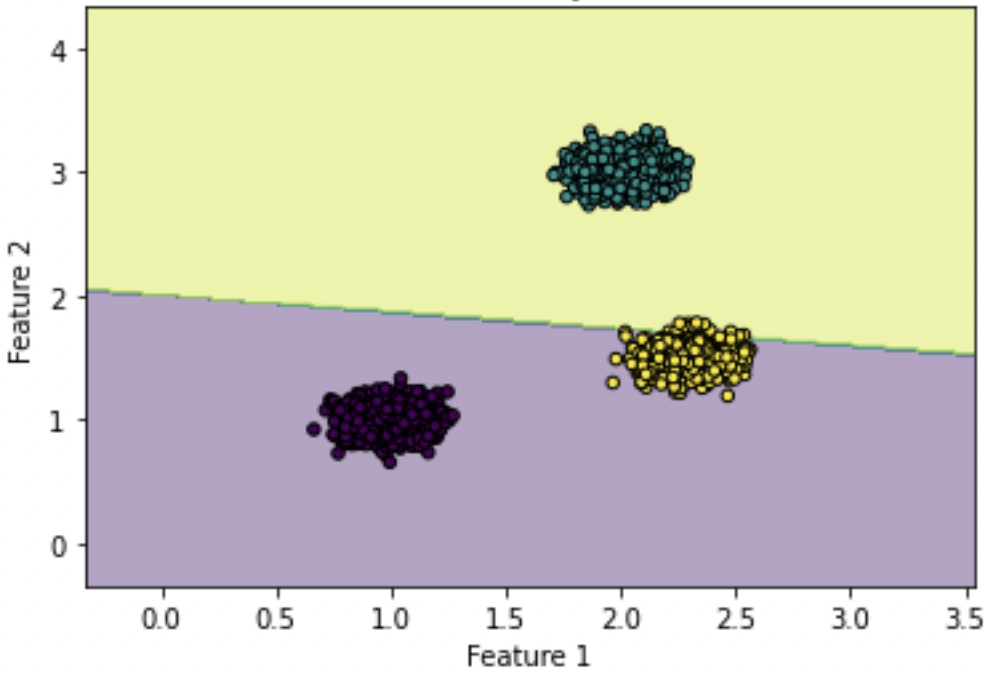}\hspace{25pt}
        \includegraphics[width=0.3\textwidth]{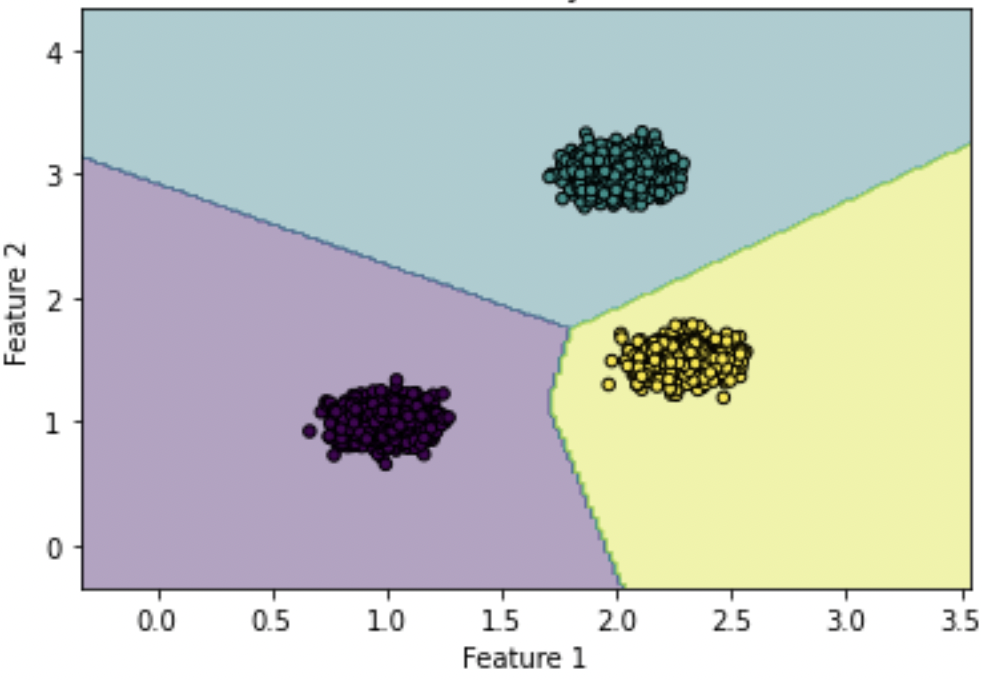}
        \caption{$3$ users with datasets from $3$ different classes each.}
        \label{fig:3 users with 3 classes}
        \vspace{2pt}
     \end{subfigure}
     \hfill
     \begin{subfigure}[b]{0.45\textwidth}
         \centering
        \includegraphics[width=0.3\textwidth]{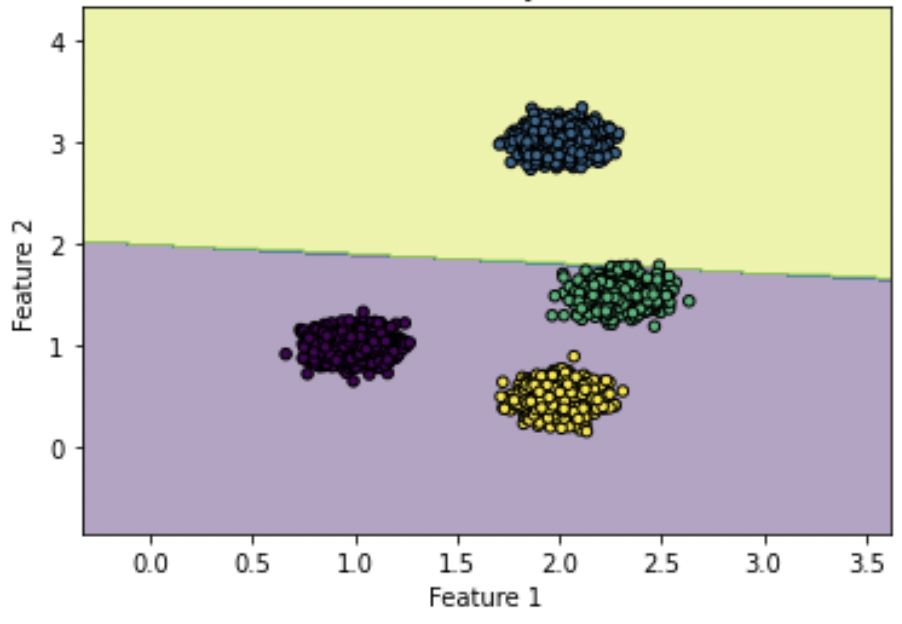}\hspace{25pt}
        \includegraphics[width=0.3\textwidth]{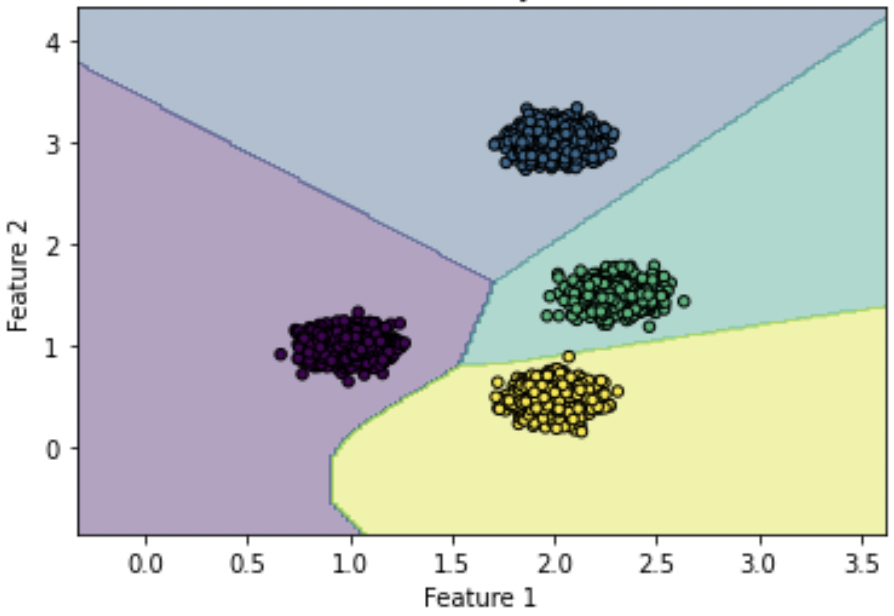}
        \caption{$3$ users with datasets of 4 classes.}
        \label{fig:3 users with 4 classes}
        \vspace{2pt}
     \end{subfigure}
     \hfill
     \begin{subfigure}[b]{0.45\textwidth}
         \centering
        \includegraphics[width=0.3\textwidth]{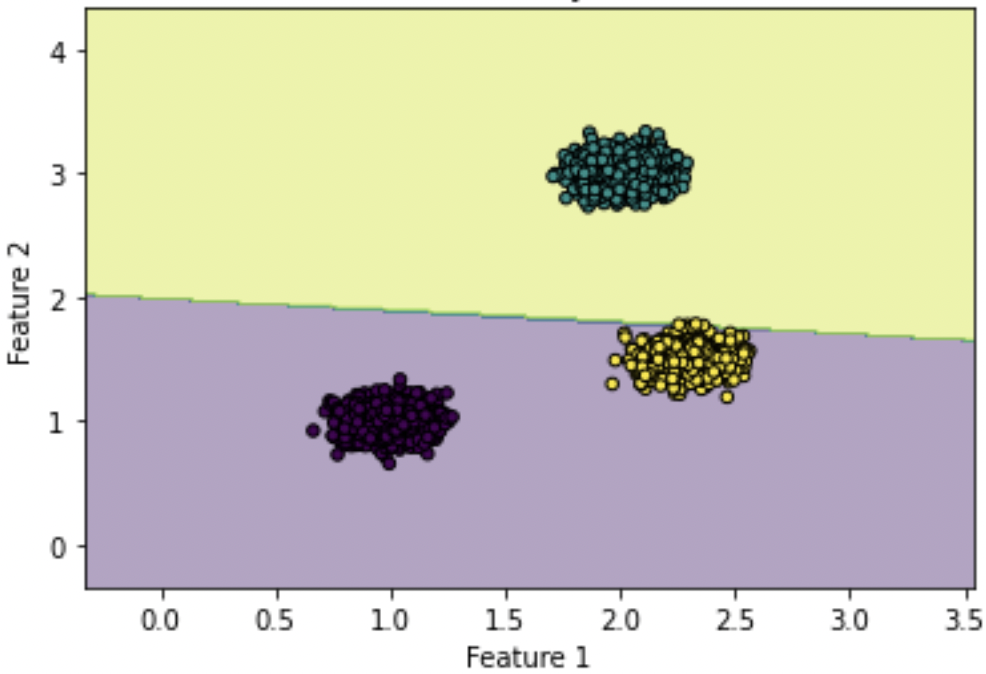}\hspace{5pt}
        \includegraphics[width=0.3\textwidth]{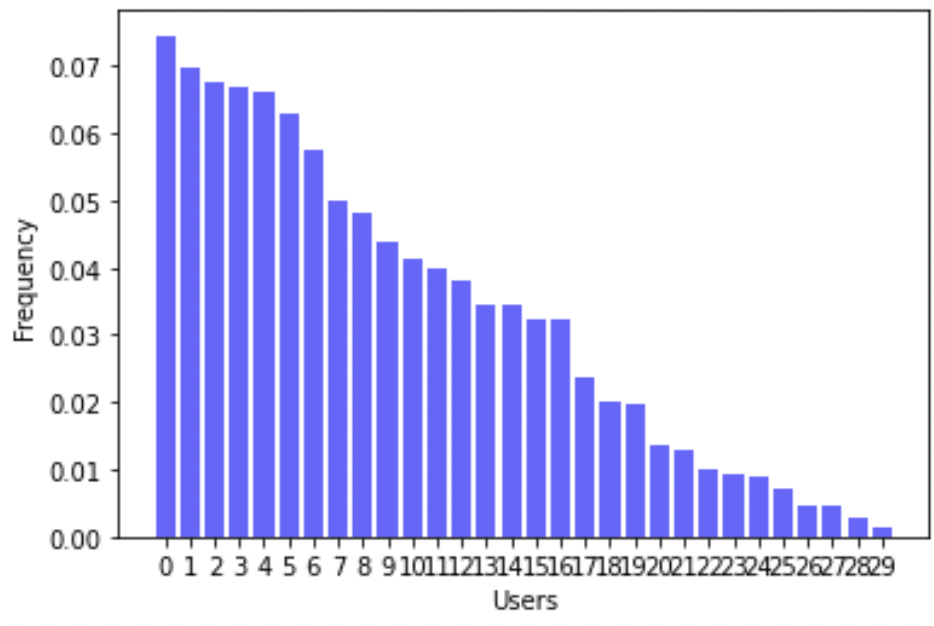}\hspace{5pt}
        \includegraphics[width=0.3\textwidth]{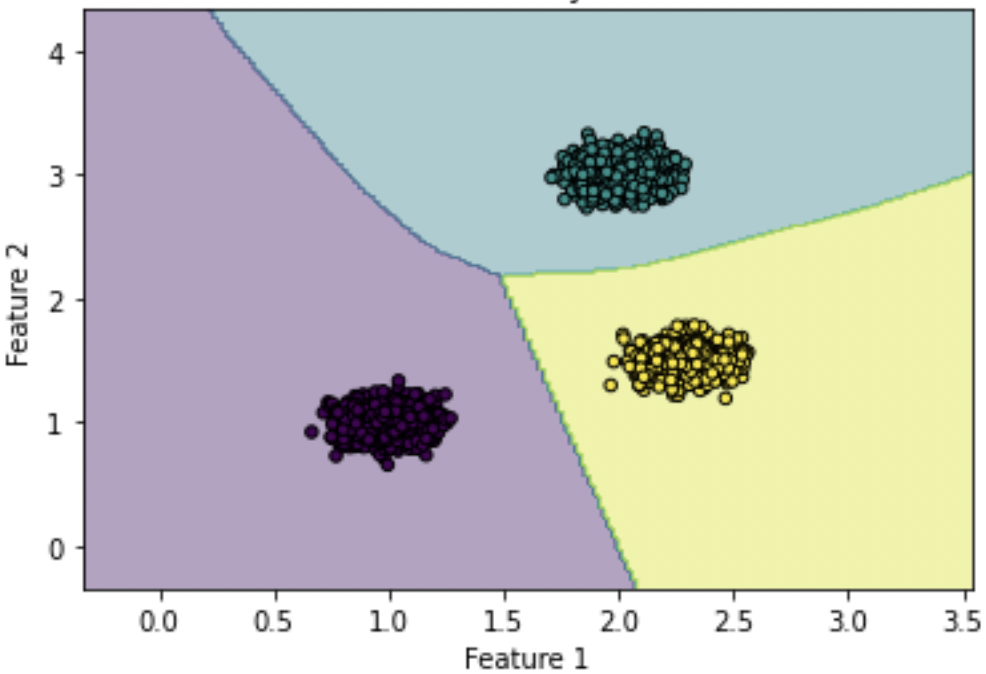}
        \caption{$30$ users with datasets of $3$ classes.}
        \label{fig:30 users with 3 classes}
        \vspace{4pt}
     \end{subfigure}
    \caption{Efficacy of our approach on a synthetic logistic regression example (Left: FL | Right: Proposed | Center: RAM Distribution).}
    \label{fig:Simple Logistic Regression Example for 2D data}
\end{figure}

\par We consider a multi-class classification setup, with features $x\in \mathcal{X} \subseteq \Re^d $ and labels $ y \in \mathcal{Y} = \{ 1,2, \cdots , C\}$, where $C $ is the total number of classes (patterns). We also consider a federated learning setting, with users $i \in [1, \dots , K] $ selected by the RAM from a fixed distribution $\mathcal{D}$, with probabilities $\rho_i, \text{such that } \sum_{i=1}^{K} \rho_i =1$. Further, we assume that each user $i$, has a private dataset $D_{i}(x,y) := D_{i}(\xi)$, with $N_i$ number of data. In the context of FL, the server updates the users via an aggregation step. Let us also define the family $ \tilde{\Theta} := \{ \phi : \mathcal{X} \times \Theta \rightarrow \Re^{C} \} $ of parameterized predictors with parameter $\theta \in \Theta \subseteq \Re^{Q} $. Each user tries to minimize its local loss function $l_{i} : \Re^{C} \times \mathcal{Y} \rightarrow \Re_{+} $.

In standard FL (and under our setting), the goal of the server would be to find optimal global parameters $\theta$ solving the problem
\begin{equation}
\begin{aligned}
\label{problem: classic FL problem}
    \inf_{\theta \in \Theta } \,   \mathbb{E}_{I \sim \mathcal{D} } \Big [ \mathbb{E} _{\xi \sim D_{I}}  \big[ 
    l_{I} ( \phi( \xi ; \theta) ) \big]   
    \Big ] ,
\end{aligned}
\end{equation}
whose  empirical version reads as
\begin{equation}
\begin{aligned}
\label{problem: classic FL problem empirical version}
    \inf_{\theta \in \Theta } \, \bigg\{ {G}_{\text{FL}}(\theta) := \sum_{i=1}^{K} \rho_{i} 
    \frac{1}{N_{i}}\sum_{j=1}^{N_{i}} 
    l_{i} ( \phi( \xi_{j} ; \theta) ) = \sum_{i=1}^{K} \rho_{i} 
    f_{i} (\theta)\bigg\}.
\end{aligned}
\end{equation}
\par Recall that the RAM applies some user selection policy which is \textit{both unknown and untouchable} from the side of the server. Therefore, any training algorithm should be agnostic to the RAM distribution, and it should work for any RAM \textit{and} dataset distributions. If the RAM distribution is highly skewed, then problem \eqref{problem: classic FL problem empirical version} faces the issue of non-uniform and rare user participation. In such a case the data are distributed with strong heterogeneity among users, data starvation exists, and accurate classification by solving \eqref{problem: classic FL problem empirical version} is generally hopeless, as we demonstrate in Fig. \ref{fig:Simple Logistic Regression Example for 2D data}(left). 
\vspace{-10pt}
\section{Proposed Approach}\label{subsec: Algorithm} \vspace{-8pt}

To guarantee efficient classification and simultaneously mitigate the effect of data starvation, we propose a \textit{risk-aware objective} that combines the \textit{Conditional Value-at-Risk} (CVaR) on the RAM distribution with the empirical risk-neutral objective \eqref{problem: classic FL problem empirical version}. The CVaR of a random variable $Z$ at confidence level $a \in (0,1]$ is defined as \cite{SIAM:Shapiro_etal} 
\begin{equation}
\begin{aligned}
\label{CVaR Definition} 
    \cvar \big[ Z \big] := \inf_{t \in \Re} \Big\{ t + \frac{1}{\alpha} \mathbb{E} \big[ ( Z - t )_{+} \big]  \Big\},
\end{aligned}
\end{equation}
for which it is true that $\cvar \big[ Z \big]=\mathbb{E}[Z]$ for $\alpha =1$, rising up to $\ess  Z$, as $\alpha\rightarrow 0$.
Then, for $\gamma\in (0,1)$, our proposed optimization problem is
\begin{align} 
\begin{aligned} 
\label{problem: CVaR FL problem}
\hspace{-3bp}
    \inf_{\theta \in \Theta } &\,
        \Big\{G^\text{R}_{\text{FL}}\big(\theta) \hspace{-1bp} := \hspace{-1bp} (1-\gamma) \cvar_{I \sim \mathcal{D}} \big[ f_{I}(\theta) \big] + 
        \gamma \mathbb{E}_{ I \sim \mathcal{D} } \big[ f_{I}(\theta) \big] \Big\},\hspace{-12bp}
\end{aligned}
\end{align}
and using definition \eqref{CVaR Definition} for $Z=f_I(\theta)$, we have
\begin{align} 
\begin{aligned} 
\label{problem: CVaR FL problem I}
   \hspace{-1pt} \inf_{\theta \in \Theta } &\,
        (1-\gamma) \inf_{t \in \Re} \bigg\{ t + \frac{1}{\alpha} \sum_{i=1}^{K} \rho_i \big[ f_{i}(\theta) - t \big]_{+} \bigg\}  + \gamma \sum_{i=1}^{K} \rho_i f_i (\theta).\hspace{-11pt}
\end{aligned}
\end{align}
The objective in the preceding problem may be further simplified as
\begin{align} 
\begin{aligned} 
\label{problem: CVaR FL problem II}
    \hspace{-4pt}
    \inf_{(\theta,t)  }   & \sum_{i=1}^{K} \rho_i  
    \bigg\{ G_{i} (\theta , t) \hspace{-2pt} := \hspace{-2pt}
    (1\hspace{-1pt}-\hspace{-1pt}\gamma) \Big[ t \hspace{-1pt}+\hspace{-1pt} \frac{1}{\alpha} \big[ f_{i}(\theta) \hspace{-1pt} - t \big]_{+} \Big]
              \hspace{-1pt}+\hspace{-1pt}  \gamma f_i (\theta)
    \bigg\}.\hspace{-12pt}
\end{aligned}
\end{align}

The CVaR measures expected losses restricted to the upper tail of the distribution of the random variable $Z$ \cite{kalogerias2020noisy}. Thus, by tuning the parameters $\gamma \in [0,1] $ and $\alpha \in (0,1] $, we tune the objective in 
\eqref{problem: CVaR FL problem} to boost the learning procedure on data points that come from rare user participation events, and essentially enforce learning under data starvation with a few shots only. Equivalently, a training algorithm based on \eqref{problem: CVaR FL problem} learns how to reject samples from frequent users since $\cvar[\cdot]$ is robust to the uncertainty of the environment \cite{SIAM:Shapiro_etal} .

\par Problem \eqref{problem: CVaR FL problem II}  leads us to devise Algorithm \ref{Algorithm RAM FL} for tackling \eqref{problem: CVaR FL problem}. Algorithm \ref{Algorithm RAM FL} is an extension of FedAvg, and essentially an instance of FedAvg on the proposed risk-aware problem \eqref{problem: CVaR FL problem II}. In each round, the server receives the parameters $(\theta_{i}^{n},t_{i}^{n})$ of a certain user $i$ (chosen iid by the RAM and not by the server) and broadcasts those as global parameters 
$(\theta_{\text{global}}^{n} , t_{\text{global}}^{n} )$ to all users. Then, each user locally applies gradient descent steps on its private dataset, updating its parameters  $\theta_{i}^{n+1} $ and $t_{i}^{n+1} $. 

We again note that the server is agnostic to RAM user selection. So, the proposed Algorithm \ref{Algorithm RAM FL} asks the users to tackle a more general problem than in the standard risk-neutral case (cf. \eqref{problem: classic FL problem}), to solve locally. Indeed, Algorithm \ref{Algorithm RAM FL} asks the users to optimize the risk-aware objective of \eqref{problem: CVaR FL problem} --through that of \eqref{problem: CVaR FL problem II}--, given a desired CVaR confidence level $a$ and a trade-off parameter $\gamma$. 
When $\alpha=1$, \eqref{problem: CVaR FL problem} is reduced to the standard FL objective \eqref{problem: classic FL problem}, and Algorithm \ref{Algorithm RAM FL} reduces to standard FedAvg. 
    
                    
                

\begin{algorithm}[!t]
\caption{FED-CVaR-AVG }
\label{Algorithm RAM FL}
\textbf{Initialize} $\theta_{i}^{1}=\theta$, $t_{i}^{1}=t$, for all $i$.
\textbf{Set} $K$, $T$, $H$, $\gamma$, $\alpha$.
\begin{algorithmic}[1]
\For{each global round $n =1,\dots,T$}

    \State{Server broadcasts the user selected by the \textbf{RAM}:
    \vspace{-0.23cm}
    \begin{equation*}
        (\theta^{n}_{\text{global}}, 
        t^{n}_{\text{global}} )
        \gets \Call{\textbf{RAM}}{ \{ (\theta^{n}_{i}, t^{n}_{i}) \}_{i=1}^{K} } 
    \end{equation*} }
    \vspace{-0.55cm}
    \For{all users $i \in [K]$ in parallel }
    
        \State{$\mathcal{B} \gets \textit{split each } D_{i} \textit{ into batches of size  }B_i $}
        
        \State{$(\theta_{i}, t_{i}) \gets (\theta^{n}_{\text{global}}, t^{n}_{\text{global}})$}
        \For{local epoch $h = 1, \dots, H$ }
            \For{batch $b \in \mathcal{B}$}
                \vspace{2pt}
                \State{$ 
                \begin{bmatrix} \theta_{i} \\ t_{i}
                \end{bmatrix}
                \gets 
                \begin{bmatrix} \theta_{i} - \eta_{\theta}  
                    \nabla_{\theta} G_{i}                   (b;\theta_{i},t_{i} ) \\ 
                    t_{i} - \eta_{t}  
                    \nabla_{t} G_{i}                    (b;\theta_{i},t_{i} )
                \end{bmatrix}
            $}
            \vspace{2pt}
            \EndFor 
        \EndFor
        \State{Forward to \textbf{RAM}: $(\theta_i^{n+1},t_i^{n+1}) \gets (\theta_i,t_i)$}
    \EndFor
\EndFor
\end{algorithmic}
\end{algorithm}

\begin{table*}[!t]
\centering
\begin{adjustbox}{width=\textwidth}
\begin{tabular}{|l|c|c|c|c|} 
\hline
 &
  \cellcolor[HTML]{38FFF8}{ $\boldsymbol{\alpha = 1.0 }$ } &
  $\boldsymbol{\alpha = 0.3}$ &
  $\boldsymbol{\alpha = 0.2 }$ &
  $\boldsymbol{\alpha = 0.1}$ \\ \hline
\multicolumn{1}{|l|}{} &
  \textbf{Overall. | pattern 1. | pattern 2.} &
  \textbf{Overall. | pattern 1. | pattern 2.} &
  \textbf{Overall. | pattern 1. | pattern 2.} &
  \textbf{Overall. | pattern 1. | pattern 2.} \\ \hline
$\boldsymbol{\gamma = 0.0 }$ &
  \cellcolor[HTML]{38FFF8}{ 
  \normalsize
  $85.147$ \tiny$ \pm 0.546$ 
  \normalsize\textbf{|}
  $85.473$ \tiny$ \pm 1.548$ 
  \normalsize\textbf{|}
  $61.041$ \tiny$ \pm 3.554$ } &
  { 
  \normalsize
  $86.157$ \tiny$ \pm 0.559$ 
  \normalsize\textbf{|}
  $88.323$ \tiny$ \pm 1.140$ 
  \normalsize\textbf{|}
  $67.164$ \tiny$ \pm 3.344$ 
  } &
  { 
  \normalsize
  $86.453$ \tiny$ \pm 0.592$ 
  \normalsize\textbf{|}
  \color{green}$\boldsymbol{88.736}$ \tiny$ \boldsymbol{ \pm 1.472 }$\color{black}
  \normalsize\textbf{|}
  $68.926$ \tiny$ \pm 2.716$ 
  } &
  \cellcolor[HTML]{FE996B}{ 
  \normalsize
  $86.089$ \tiny$ \pm 0.923$ 
  \normalsize\textbf{|}
  $87.536$ \tiny$ \pm 2.449$ 
  \normalsize\textbf{|}
  $70.601$ \tiny$ \pm 3.523$ 
  } \\ \hline
$\boldsymbol{\gamma = 0.1 }$ &
  \cellcolor[HTML]{38FFF8}{ 
  \normalsize
  $84.895$ \tiny$ \pm 0.605$ 
  \normalsize\textbf{|}
  $84.583$ \tiny$ \pm 1.597$ 
  \normalsize\textbf{|}
  $59.649$ \tiny$ \pm 3.795$ 
  } &
  { 
  \normalsize
  $86.276$ \tiny$ \pm 0.519$ 
  \normalsize\textbf{|}
  $88.353$ \tiny$ \pm 1.058$ 
  \normalsize\textbf{|}
  $66.676$ \tiny$ \pm 2.883$ 
  } &
  { 
  \normalsize
  $86.432$ \tiny$ \pm 0.653$ 
  \normalsize\textbf{|}
  $88.619$ \tiny$ \pm 1.674$ 
  \normalsize\textbf{|}
  $68.801$ \tiny$ \pm 2.835$ 
  } &
  \cellcolor[HTML]{FE996B}{ 
  \normalsize
  \color{green}$\boldsymbol{86.546}$ \tiny$ \boldsymbol{ \pm 0.540 }$ \color{black}
  \normalsize\textbf{|}
  $88.681$ \tiny$ \pm 1.454$ 
  \normalsize\textbf{|}
  \color{green}$ \boldsymbol{ 71.681 }$ \tiny$ \boldsymbol{ \pm 2.419 }$ \color{black}
  } \\ \hline
$\boldsymbol{\gamma = 0.2 }$ &
  \cellcolor[HTML]{38FFF8}{
  \normalsize
  $84.783$ \tiny$ \pm 0.474$ 
  \normalsize\textbf{|}
  $84.533$ \tiny$ \pm 1.294$ 
  \normalsize\textbf{|}
  $59.223$ \tiny$ \pm 2.580$ 
  } &
  { 
  \normalsize
  $86.174$ \tiny$  \pm 0.552$ 
  \normalsize\textbf{|}
  $88.029$ \tiny$  \pm 1.572$ 
  \normalsize\textbf{|}
  $66.372$ \tiny$  \pm 3.116$ 
  } &
  {
  \normalsize
  $86.110$ \tiny$ \pm 0.573$ 
  \normalsize\textbf{|}
  $88.061$ \tiny$ \pm 1.663$ 
  \normalsize\textbf{|}
  $67.659$ \tiny$ \pm 3.156$ 
  } &
  \cellcolor[HTML]{FE996B}{ 
  \normalsize
  $86.132$ \tiny$ \pm 0.860$ 
  \normalsize\textbf{|}
  $88.093$ \tiny$ \pm 1.990$ 
  \normalsize\textbf{|}
  $69.698$ \tiny$ \pm 4.351$ 
  } \\ \hline
$\boldsymbol{\gamma = 0.3 }$ &
  \cellcolor[HTML]{38FFF8}{
  \normalsize
  $84.983$ \tiny$ \pm 0.419$ 
  \normalsize\textbf{|}
  $84.969$ \tiny$ \pm 1.042$ 
  \normalsize\textbf{|}
  $60.330$ \tiny$ \pm 2.104$ 
  } &
  { 
  \normalsize
  $86.036$ \tiny$ \pm 0.483$ 
  \normalsize\textbf{|}
  $87.611$ \tiny$ \pm 1.316$ 
  \normalsize\textbf{|}
  $65.958$ \tiny$ \pm 2.492$ 
  } &
  {
  \normalsize
  $86.263$ \tiny$ \pm 0.517$ 
  \normalsize\textbf{|}
  $88.300$ \tiny$ \pm 1.772$ 
  \normalsize\textbf{|}
  $68.076$ \tiny$ \pm 3.054$ 
  } &
  \cellcolor[HTML]{FE996B}{ 
  \normalsize
  $86.197$ \tiny$ \pm 0.533$ 
  \normalsize\textbf{|}
  $87.776$ \tiny$ \pm 2.026$ 
  \normalsize\textbf{|}
  $69.883$ \tiny$ \pm 2.674$ 
  } \\ \hline
\multicolumn{1}{|l|}{\cellcolor[HTML]{38FFF8}{$\boldsymbol{\gamma = 1.0} $} } 

  & \cellcolor[HTML]{38FFF8}{
  \normalsize
  $84.915$ \tiny$ \pm 0.497$ 
  \normalsize\textbf{|}
  $84.514$ \tiny$ \pm 1.620$ 
  \normalsize\textbf{|}
  $60.252$ \tiny$ \pm 3.074$ 
  } 
  
  & \cellcolor[HTML]{38FFF8}{
  \normalsize
  $84.822$ \tiny$ \pm 0.630$ 
  \normalsize\textbf{|}
  $84.481$ \tiny$ \pm 2.003$ 
  \normalsize\textbf{|}
  $59.008$ \tiny$ \pm 3.138$ 
  } 
  
  & \cellcolor[HTML]{38FFF8}{
  \normalsize 
  $84.916$ \tiny$ \pm 0.454$ 
  \normalsize\textbf{|} 
  $85.459$ \tiny$ \pm 2.870$ 
  \normalsize\textbf{| }
  $60.805$ \tiny$ \pm 2.399$ 
  } 
  
  & \cellcolor[HTML]{38FFF8}{
  \normalsize 
  $84.956$ \tiny$ \pm 0.462 $ 
  \normalsize\textbf{|} 
  $85.561$ \tiny$ \pm 1.443 $ 
  \normalsize\textbf{|} 
  $59.839$ \tiny$ \pm 3.168 $
  } 
 \\ \hline
\end{tabular}
\end{adjustbox}
\vspace{5bp}
\caption{ \textbf{FashionMnist:} The $10\%$ of less available users (that means $3$ out of $30$ users with sampling probabilities $ 0.0107, 0.0078, 0.0053 $, respectively) carry $2$ patterns exclusively. The columns represent the overall testing accuracy, and the testing accuracy, of the global model, at the patterns that belong to the less available users, respectively. The objective becomes risk-neutral when $\alpha =1.0$, or $\gamma=1.0$.}
\vspace{-10bp}
\label{Table:FashionMnist 2 classes}
\end{table*}

\vspace{-9pt}
\subsection{A Motivating Example } \label{Simple motivating example}
\vspace{-4pt}

We now present a simple example to illustrate the differences in behavior of the standard (risk-neutral) FL objective in \eqref{problem: classic FL problem} and the proposed risk-aware objective in \eqref{problem: CVaR FL problem}. Suppose that a dataset is distributed among $K=3$ users, with each user training for $1$ pattern. Let us also assume that the RAM selects the users with probabilities $\rho_1 > \rho_2 \gg \rho_3$, which means that the RAM allows users $1$, and $2$, to communicate more often with the server than the user $3$.

As usual, the classical FedAvg \cite{mcmahan2017communication} approach will try to minimize the objective function 

\vspace{-10pt}
\begin{equation}
\begin{aligned}
\label{motivating example: classic FL version}
    {G}_{\text{FL}}(\theta) =& \sum_{i=1}^{K=3} \rho_{i}f_{i}(\theta) 
     = \rho_{1}f_{1}(\theta) + \rho_{2}f_{2}(\theta) + \rho_{3}f_{3}(\theta),
\end{aligned}
\end{equation}
where the weights $\rho_1,\rho_2$ and $\rho_3$ are unknown, but implicitly supplied by the RAM.

On the other hand, for a sufficiently small and strictly positive choice of the hyper-parameter (the CVaR level) $\alpha$, it can be easily shown (although not entirely trivially) that the positive part of the risk-aware objective in  \eqref{problem: CVaR FL problem} is activated only for the upper $\alpha$-quantile of the empirical losses $f_1(\theta),f_2(\theta),f_3(\theta)$ on the random variable $I$ (for each fixed $\theta$). This yields the \textit{weighted user-robust loss}
\begin{equation}
\label{motivating example: CVaR version part 2}
    G^\text{R}_{\text{FL}}\big(\theta) 
    = 
    (1 - \gamma) \max\{f_1(\theta),f_2(\theta),f_3(\theta)\}
    +
    \gamma {G}_{\text{FL}}(\theta),
\end{equation}
for every sufficiently small trade-off parameter $\gamma\in[0,1]$.
We observe that the risk-aware objective \eqref{motivating example: CVaR version part 2} focuses on the worst user loss regardless of the corresponding probability of it being selected by the RAM, with relative proportion $1-\gamma$. 

Therefore, in a region of the space where, e.g., $f_3$ is larger, which is expected to happen due to rare sampling by the RAM, the risk-aware objective \eqref{motivating example: CVaR version part 2} will steer $\theta$ towards regions of $\Theta$ that equalize (i.e., reduce) the values of the local training loss $f_3$, relative to $f_1$ and $f_2$. In other words, the objective \eqref{motivating example: CVaR version part 2} induces \textit{user equity} in FL, which is initially hindered by the presence of the RAM.
\begin{figure*}[!h]
    \hspace{-2pt}
    \begin{subfigure}[b]{0.37\textwidth}
        \centering
        \includegraphics[width=\textwidth, height=0.25\textwidth]{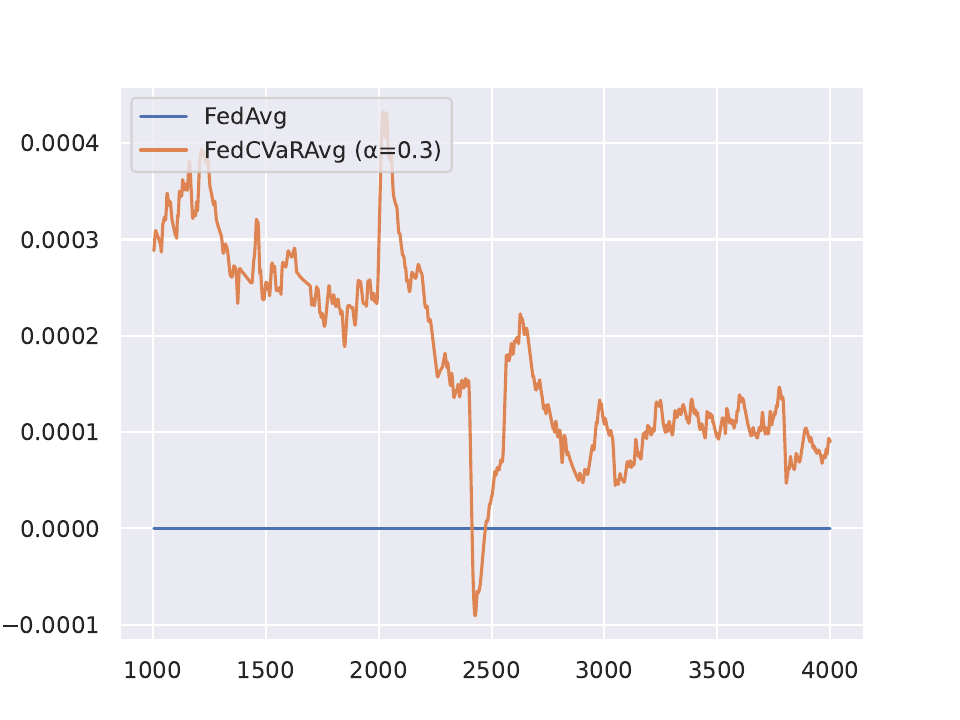}
        \caption{ Global CVaR Threshold}
        \label{fig:Global t Mnist}
    \end{subfigure}
    \hspace{-23pt}
    \begin{subfigure}[b]{0.37\textwidth}
        \centering
        \includegraphics[width=\textwidth, height=0.25\textwidth]{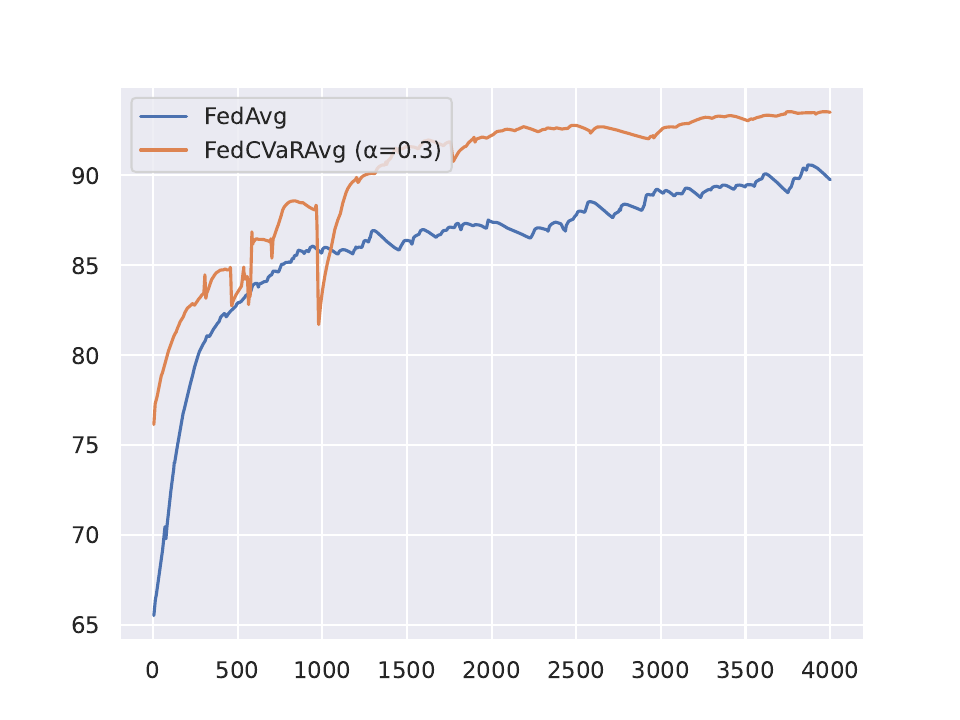}
        \caption{ Overall Test Accuracy }
        \label{fig:Overall Test Accuracy}
    \end{subfigure}
    \hspace{-23pt}
        \begin{subfigure}[b]{0.37\textwidth}
        \centering
        \includegraphics[width=\textwidth, height=0.25\textwidth]{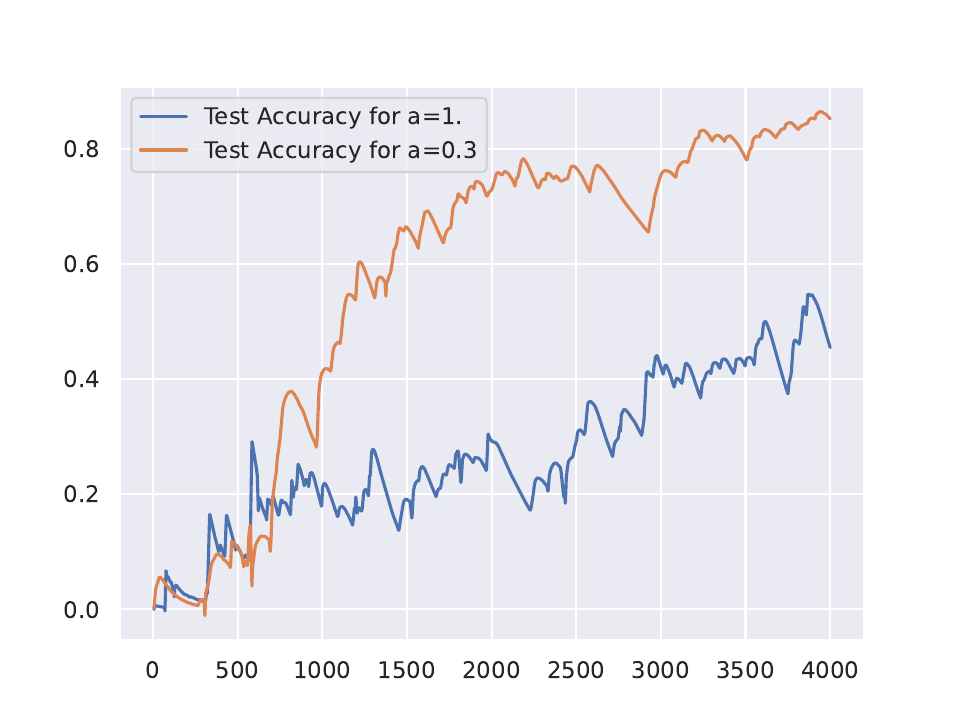}
        \caption{ Test Accuracy for Infrequent Classes}
        \label{fig:Test Accuracy at class of less often users}
    \label{fig:Mnist 1 class}
    \end{subfigure}
    \hfill
    \vspace{-8pt}
    \caption{\textbf{MNIST}: For $K=30$ users, with the $3$ of less often users have exclusively $1$ of the patterns.}  
    \label{fig:MNIST 1}
    \vspace{-14pt}
\end{figure*}

It is worth-noting that
while \eqref{motivating example: CVaR version part 2} is an operationally desirable objective,
it is practically impossible for the server to infer which of the three losses is largest, since the RAM prevents the server from controlling user participation in learning. Additionally, \eqref{motivating example: CVaR version part 2} generally results in not well-behaved and possibly nonsmooth FL problems. In our approach, these challenges are effectively addressed by replacing the risk-aware problem \eqref{problem: CVaR FL problem} by its equivalent version \eqref{problem: CVaR FL problem II}, which is well-behaved and does not require access to unavailable information.


The sharp distinction between \eqref{motivating example: classic FL version} and \eqref{motivating example: CVaR version part 2} is demonstrated in Fig. \ref{fig:Simple Logistic Regression Example for 2D data}, for a simple logistic regression setup on 2D synthetic data. Specifically, in Fig. \ref{fig:3 users with 3 classes}, each of $K=3$ users receives one pattern, while in \ref{fig:3 users with 4 classes}, an extra pattern is assigned to the least frequent user $3$.
User selection probabilities are $\rho_{1} = 0.5, \rho_{2}=0.4,$ and, $ \rho_{3}=0.1 $. We observe that the standard FL ($\alpha=1$) fails to classify correctly the pattern/s from the least frequent user $3$. However, the proposed approach which solves \eqref{problem: CVaR FL problem II} (for $\alpha=0.1$) generates decision boundaries that correctly classify all patterns in both cases. Similarly, in Fig. \ref{fig:30 users with 3 classes}, we scale up to $K=30$ users, with the $6$ least frequent users training on the yellow dataset. The user selection probabilities are shown in \ref{fig:30 users with 3 classes}(center). Once again, the risk-aware approach ($\alpha=0.1$) successfully generates a decision boundary that classifies all patterns. Across all examples shown in figure \ref{fig:Simple Logistic Regression Example for 2D data}, the trade-off $\gamma$ remains constant at $0.1$ and all experiments were performed fairly under exactly the same choices regarding algorithm hyperparameters, number of epochs (all models are ``trained to plateau"), etc.

\vspace{-10pt}
\section{Experimental Results}\label{Experimental Results}
\vspace{-8.5pt}

\begin{figure}[!t]
    \centering
    \begin{subfigure}[b]{0.25\textwidth}
        \centering
        \includegraphics[width=\textwidth, height=0.25\textwidth]{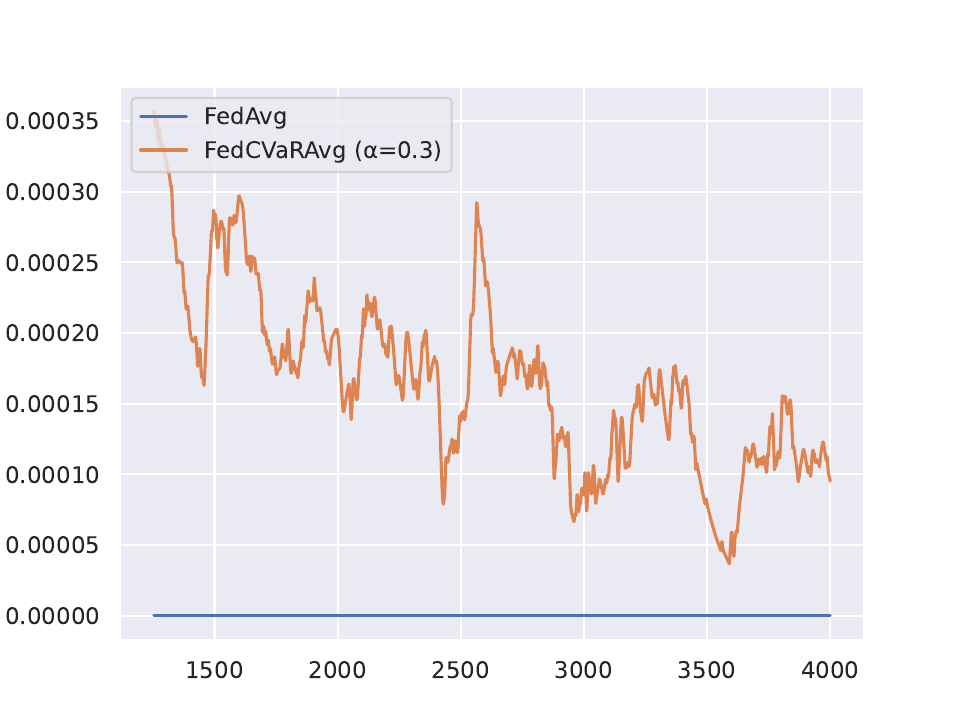}
        \caption{ Global t }
        \label{fig:Global t}
    \end{subfigure}
    \hspace{-18pt}
    \begin{subfigure}[b]{0.26\textwidth}
        \centering
        \includegraphics[width=\textwidth, height=0.25\textwidth]{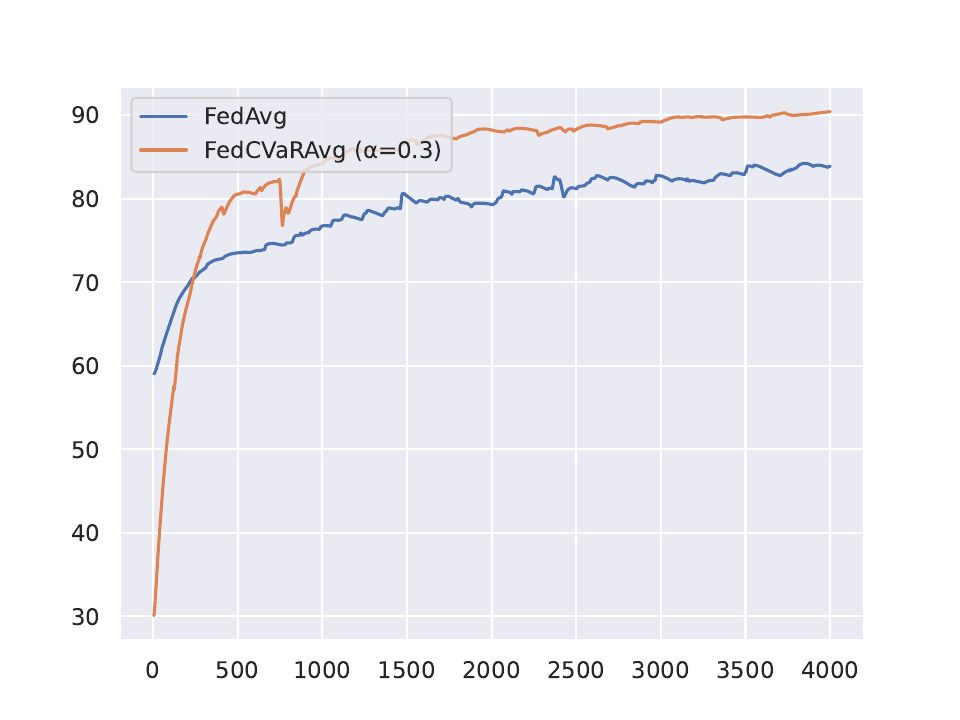}
        \caption{ Overall Test Accuracy }
        \label{fig:Overall Test Accuracy}
    \end{subfigure}
    \hspace{-30pt}
    \\ \vspace{-12pt}
    \hspace{-27.5pt}
    \begin{subfigure}[b]{0.26\textwidth}
        \centering
        \includegraphics[width=\textwidth, height=0.25\textwidth]{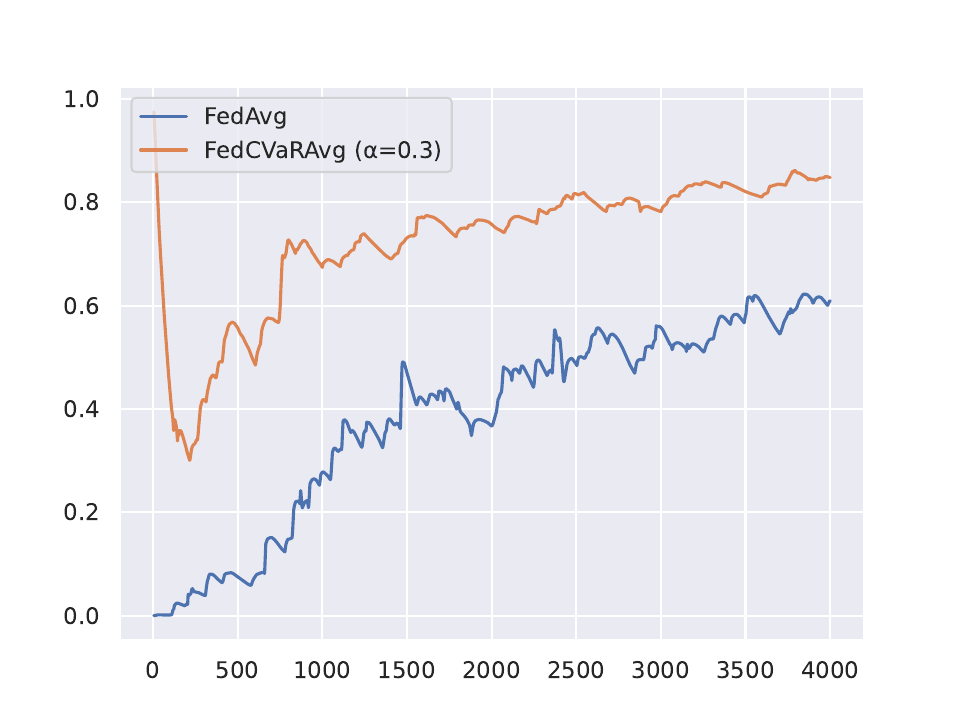}
        \caption{ Test Accuracy: Rare Class 1}
        \label{fig:Test Accuracy at 1 class of less often users}
    \end{subfigure}
    \hspace{-18pt}
    \begin{subfigure}[b]{0.26\textwidth}
        \centering
        \includegraphics[width=\textwidth, height=0.25\textwidth]{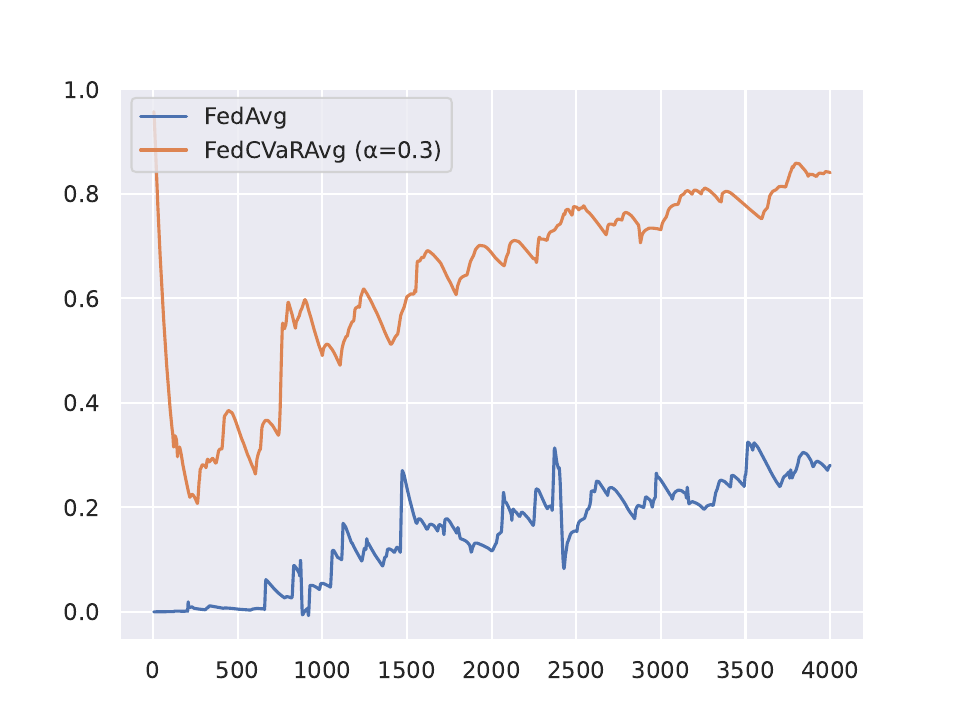}
        \caption{ Test Accuracy: Rare Class 2}
        \label{fig:Test Accuracy at 2 class of less often users}
    \end{subfigure}
    \hspace{-25pt}
    \vspace{4pt}
    \caption{\textbf{MNIST}: For $K=30$ users, with $3$ of less often users have exclusively the $2$ of the patterns.}  
\label{fig:MNIST 2}
\end{figure}

We now evaluate the proposed Algorithm \ref{Algorithm RAM FL} on the \textbf{Mnist} (Figs. \ref{fig:MNIST 1} and \ref{fig:MNIST 2}) and \textbf{FashionMnist} (Table \ref{Table:FashionMnist 2 classes}) benchmarks, each comprised of 60,000 training samples with 10 distinct patterns. Data are distributed among K users in a heterogeneous way. We split the data, reserving $M\%$ of the most frequent users for $r\%$ of available patterns, while the remaining $(100-r)\%$ of the data is uniformly distributed among the least frequent users, comprising the remaining $(100-M)\%$. The total number of global rounds is the same for both FedAvg and the proposed algorithm, and set to $4000$ and $6000$ for the \textbf{Mnist} and \textbf{FashionMnist} datasets, respectively. Code for all experiments is available at~\cite{simulations}.



For the \textbf{Mnist} dataset, we present results for two experiments with $K=30$ users. For the first experiment in Fig. \ref{fig:MNIST 1}, we set $M=90\%$, and $ r=90\% $, and for the second experiment in Fig. \ref{fig:MNIST 2}, we set 
$M=90\%$, and $ r=80\% $. We also choose $\alpha = 0.3$ and $\gamma = 0.3 $. For both FedAvg and Algorithm \ref{Algorithm RAM FL} we use a neural network with two fully-connected hidden layers, with number of neurons $(128,128)$ \cite{shen2021agnostic}. Stepsizes are set constant as $\eta_{\theta} = 10^{-3}$, $\eta_{t} = 10^{-4}$, and each user conducts $10$ local epochs. We report smoothed graphs for clarity. In both Figs. \ref{fig:MNIST 1} and \ref{fig:MNIST 2}, we readily observe that Algorithm \ref{Algorithm RAM FL} achieves both better overall performance and better performance at the patterns that are locally trained by the least frequent users. We can observe performance improvement to over $80\%$, from around $50\%$ (Fig. \ref{fig:Test Accuracy at class of less often users}), $60\%$ (Fig. \ref{fig:Test Accuracy at 1 class of less often users}) and $30\%$ (Fig. \ref{fig:Test Accuracy at 2 class of less often users}), respectively.


For the \textbf{FashionMnist} dataset, Table \ref{Table:FashionMnist 2 classes} provides a more detailed range of experiments. We set $K=30$, $M=90\%$ and $ r=80\% $, and use a CNN similar as in \cite{mcmahan2017communication} with two $5 \times 5$ convolutional layers (with $6$ and $16$ channels, respectively, each followed with $2 \times 2$ max pooling) and two fully connected layers with $120$ and $84$ neurons. Stepsizes are set as $\eta_{\theta} = 10^{-2}$ and $\eta_{t} = 5\times 10^{-4}$. Each user runs $20$ local epochs. The setting remains the same for all $\alpha$ and $\gamma$.

Table \ref{Table:FashionMnist 2 classes} shows the overall global performance as well as that for the patterns belonging to the least frequent users, for a variety of pairs $( \gamma, \alpha )$. The blue area of Table \ref{Table:FashionMnist 2 classes} corresponds to cases where problem \eqref{problem: CVaR FL problem} is essentially reduced to standard FL. On the other hand, in the orange area the objective of \eqref{problem: CVaR FL problem} becomes most risk-sensitive.
We observe that the accuracy at pattern $2$ has been improved from approximately $60 \%$ for standard FL to more than $71 \%$, for $\alpha = 0.1$ and $\gamma = 0.1$. Moreover, performance on pattern $1$ is also improved as compared with FedAvg. Further, it is worth mentioning that pattern $2$ is most difficult to learn even for the standard FL, compared with pattern $1$. As expected, overall performance improves when $\alpha \rightarrow 0$, with the case $(\gamma = 0.1, \alpha = 0.1)$ performing the best. 



\vspace{-6pt}
\section{Conclusion}
\vspace{-6pt}

In this work, we studied FL under an unknown random access model (RAM) describing biased, non-uniform, skewed and/or restricted random user participation. Departing from the standard expected loss model, we proposed a new risk-aware objective constructed by taking the CVaR over the RAM distribution, resulting in an efficient training algorithm which is oblivious
to the RAM, but at the same time addresses
limited participation of infrequent users.
Through experimental evaluation on 2D synthetic, Mnist, and FashionMnist datasets, we have demonstrated that the proposed risk-aware approach brings substantial potential performance gains over standard FL relying on FedAvg which, to the best of our knowledge, is currently a state-of-the-art method for handling FL problems with no server intervention on user participation.

\clearpage

\bibliographystyle{IEEEbib}
\bibliography{refs}

\end{document}